# Explainable few-shot learning workflow for detecting invasive and exotic tree species

Caroline M. Gevaert[1,*], Alexandra Aguiar Pedro[3], Ou Ku[2], Hao Cheng[1], Pranav Chandramouli[2], Farzaneh Dadrass Javan[1], Francesco Nattino[2], and Sonja Georgievska[2]

[1]University of Twente, Faculty ITC, Enschede, 7500 AE, the Netherlands
[2]Netherlands eScience Center, Amsterdam, 1098 XH, the Netherlands
[3]São Paulo Municipal Green and Environment Secretariat, São Paulo, 04103-000, Brazil
[*]c.m.gevaert@utwente.nl

## ABSTRACT

Deep Learning methods are notorious for relying on extensive labeled datasets to train and assess their performance. This can cause difficulties in practical situations where models should be trained for new applications for which very little data is available. While few-shot learning algorithms can address the first problem, they still lack sufficient explanations for the results. This research presents a workflow that tackles both challenges by proposing an explainable few-shot learning workflow for detecting invasive and exotic tree species in the Atlantic Forest of Brazil using Unmanned Aerial Vehicle (UAV) images. By integrating a Siamese network with explainable AI (XAI), the workflow enables the classification of tree species with minimal labeled data while providing visual, case-based explanations for the predictions. Results demonstrate the effectiveness of the proposed workflow in identifying new tree species, even in data-scarce conditions. With a lightweight backbone, e.g., MobileNet, it achieves a F1-score of 0.86 in 3-shot learning, outperforming a shallow CNN. A set of explanation metrics, i.e., *correctness*, *continuity*, and *contrastivity*, accompanied by visual cases, provide further insights about the prediction results. This approach opens new avenues for using AI and UAVs in forest management and biodiversity conservation, particularly concerning rare or under-studied species.

## Introduction

Earth Observation imagery, such as satellites and Unmanned Aerial Vehicles (UAVs), combined with modern Artificial Intelligence (AI) algorithms, enables rapid mapping and proves its use for numerous sustainable development goals. For example, monitoring invasive species in natural ecosystems. One significant challenge, however, is that these AI algorithms depend on the availability of large amounts of labeled data for training, which is often lacking in Low and Middle-Income Countries (LMICs). The lack of labels induces two challenges: 1) it is difficult to train Deep Learning models to the task at hand, and 2) it is difficult to evaluate these models to see if they are performing well. Researchers focus on the first challenge through the development of few-shot learning algorithms. However, the second challenge is often neglected, as few-shot learning papers use an evaluation set that still requires labeled samples to assess whether a model is performing well. If labeled data is unavailable in the field for training an algorithm, it will also not be available to assess its performance. To this end, this manuscript proposes a workflow that combines few-shot learning with explainable AI for a novel study that aims to identify invasive species in the Atlantic forests of Brazil. These species are typically identified in the field by experts walking along forest paths.

Effective, continuous monitoring of forests is crucial to manage and conserve forests efficiently and their ecosystem services from threats such as deforestation, climate change, stress and diseases, and invasive species[1]. Remote sensing offers several advantages over traditional field-based methods for monitoring forests where experts walk along forest paths to identify species, including scalability, multi-sensor capabilities, cost and time effectiveness[2]. Satellite imagery is suitable for covering large areas. However, more recently, UAVs have emerged as adaptive tools to collect higher-resolution data for monitoring forests, and they are offering promising solutions to overcoming the challenges of satellite-based data in terms of spatial and temporal resolution[3]. Both satellite and UAV imagery can be coupled with AI techniques, such as Machine Learning, to automate the extraction of information, such as identifying invasive species, from the imagery.

The added spatial detail provided by UAVs shows promising results for identifying and characterizing Atlantic rain forests. In 2022, one study manually delimited tree crowns over hyperspectral UAV imagery and then performed a random forest classification to identify eight species[4]. Another research in 2023 combined hyperspectral UAS and LiDAR to identify eight tree species[5]. Other studies used UAS-LiDAR and hyperspectral data to predict tree characteristics such as aboveground biomass, canopy height, and leaf area index[6]. Similarly, 3D characteristics from UAV-LiDAR were processed with random forests to detect *Araucaria angustifolia*[7] and distinguish between stages of forest regeneration[8]. Hyperspectral and LiDAR systems are generally more expensive, but research has used low-cost UAS that capture only RGB imagery to monitor Atlantic forests. For example, Dasilva et al. compared random forest and SVM to identify invasive species[9] and Albuquerque et al. used random forest classification to distinguish characteristics such as tree density, tree height, and vegetation cover[10]. Deep Learning methods are also gaining traction; some studies use ResNet[11] or Faster RCNN[12] to identify tree species from the RGB UAV imagery. The lack of labeled data is a significant hurdle in using AI methods for tree species detection. While many Machine Learning-based strategies depend on large quantities of well-prepared training and test samples, few-shot learning addresses scenarios where a limited number of labeled samples are available for training. In forest monitoring, few-shot learning can be beneficial. At the same time, a limited number of samples are available and can support transfer learning where we can reuse the knowledge learned from a similar task in a less accessible region. Moreover, due to the dynamic nature of the forest, few-shot learning provides rapid adoption of pre-trained models to the specific forest type, region, or situation[13].

There have been a few examples of few-shot learning in Earth Observation. One study tested Siamese networks with contrastive learning for few-shot learning[14], inspired by the work of Wang et al[15]. Another study proposes SCL-P-Net[16], which incorporates contrastive learning into prototypical networks for tree species classification from hyperspectral imagery. Despite the limited attention paid to few-shot learning for forest species detection, it is very relevant from a practical point of view. While data is more readily available in High-Income Countries (HICs), finding labeled datasets of local tree species in other regions is challenging. Therefore, it is essential to develop strategies that combine the advantages of Deep Learning and automatic object detection and classification with the capability of working with few training samples.

On the other hand, not only is it essential to develop a model with a high classification accuracy, but there is increasing awareness of the importance of developing explainable methods, which can help give the user confidence in results and encourage the faster development of accurate models. Interpretability and explainability are gaining traction in Earth Observation as well[17]. Explainability as a concept has many different meanings. Here, we follow other publications in Earth Observation, which relate explainability to the ability to understand how models derive a prediction through the lens of specific domain knowledge[18]. Activation maps are a popular explainability method amongst publications (e.g.,[19,20]). Such activation maps highlight parts of an image that have a relatively strong influence on the model decision[21]. For example, such attention maps have been used to provide interpretable visual cues for ship remote sensing image retrieval using Siamese networks[22]. However, there are many criticisms of such saliency or activation maps. Some argue that similar saliency maps are obtained for randomly initialized weights and trained models[23] or when queried for different classes[24]. Therefore, the "explanations" provided through salience maps are complex and non-self-evident.

Example-based explanations could be more similar to human reasoning. Research on human understanding implies that human decision-making tends to be contrastive and relatively simple[25] – only focusing on one or two reasonings rather than deriving complex answers. Moreover, a study of user preferences for explanations including saliency maps, SHAP, and example-based explanations demonstrated that users greatly preferred example-based explanations over Grad-CAM, a saliency map method, (90% vs 50% respectively) for image classification tasks[26]. Other user studies suggest that showing just one similar example as an explanation can have similar effectiveness as explanations that the user can interact with[27] and that example-based explanations can help users judge the correctness of classification outputs in domains that they are not familiar with[28]. Therefore, the utility of example-based explanations is grounded in the scientific theory of human decision-making and demonstrated through user studies comparing various explanations.

In the domain of Earth Observation, a group of researchers presented a method to verify the reliability of a Deep Learning model by showing a single image from the training dataset which is similar to the test sample to be classified, obtaining reasonable results[29]. From the medical domain, one study deploys an example-based explainability method by using nearest-neighbor to select the closest example from the same class and the closest example from a different class in the latent space[30]. Using examples from the same and different classes is similar to the approach presented here. Another study uses case-based explanations to explain the workings of a Siamese network in a class-to-class setup[31]. The literature above suggests that a case-based explanation is compatible with Siamese Networks in a few-shot learning classification setup and would be more meaningful to the user. As mentioned above, one limitation of many XAI studies[17] is that explanations are presented but rarely evaluated. Therefore, in this study, we will build on an existing XAI evaluation framework[32] to qualitatively and quantitatively evaluate the explanation provided.

To address all the challenges discussed above, this paper presents an innovative end-to-end workflow for a few-shot learning-based strategy demonstrated for detecting invasive and native species in the mid-Atlantic rainforest from UAV-based imagery. The proposed use of UAVs can capture these species in a spatially consistent manner from the sky, and the XAI workflow subsequently enables the automatic identification of tree species of interest. A shallow CNN-based Siamese network and a lightweight backbone network MobileNet[33] are suitable in environments with limited high-performance computing hardware. Finally, a case-based strategy for explanation generation is also proposed, and the quality of this explanation is evaluated.

## Materials and Methods

### Study-area

Most of the remaining areas of the Atlantic Forest in São Paulo, Brazil, are on the city's edges, where the five city's natural parks (integral protection conservation units aiming to preserve natural ecosystems of ecological relevance) are located. Despite being conservation areas, and thus protected from urban expansion, the remaining Atlantic Forest in these parks also suffer from the expansion of invasive exotic species over the native forest. This is indeed the case in the Bororé Natural Park, which was selected as a case study for this project. To define the actions that prevent biodiversity loss caused by invasive species, experts from the Municipal Green and Environment Secretariat (SVMA) conduct field inspections in the natural parks to elaborate biodiversity assessments of some existing plant formations and species. As natural parks are large areas, these assessments are carried out in only parts of the parks, that is, in a sample area. Therefore, there is no mapping of individual invasive trees in the municipal parks, let alone outside parks or conservation units. In the Bororé Natural Park, the SVMA assessment found 141 trees of invasive exotic species. Based on this assessment, a section of this sampling with the highest concentration of identified species was selected for planning the UAV flights. A DJI Mavic Pro with an FC220 RGB sensor was utilized. The UAV images were processed with Agisoft Metashape to produce the dense point cloud and the orthomosaic. Individuals of the *Archontophoenix cunninghamiana* (popular name: Seafórtia), an invasive Australian palm, were identified as the invasive species of interest (Species 7 and 10 below). Seafórtia reaches eight to ten meters in height in relatively homogeneous agglomerations. Animals, especially birds, used to be attracted by their red seeds, and they were disseminated around the forest areas. As an invasive species, Seafórtia competes with native species in the natural environment, and their expansion can significantly alter habitats, causing the local extinction of native species and generating other ecosystem complications. Adequate management and coping with this issue to protect the native forest demand adequate and efficient mapping of these species' individuals once they know their location, which is the first step in any action.

### Methods

The overall workflow of the proposed methodology is depicted in Figure 1. The required input for the workflow is a UAV orthomosaic of the forest where invasive or native tree species are detected. A few manually labeled bounding boxes indicating the species of interest and benchmark data of aerial imagery of tree canopies of a known species. The first processing step is an object detection model automatically extracting individual tree canopies from the UAV orthomosaic. This ensures a single tree canopy in each cutout. This objection detection can be performed manually based on expert knowledge or with Machine Learning models such as Netflora, which is applied to retrieve species 1 and 2 shown in Figure 2. These extracted tree canopies are combined with benchmark data to pre-train a Siamese network. In this study, to study the applicability of the proposed workflow, we also compare the shallow CNN-based Siamese network trained from scratch with a lightweight backbone-based (MobileNet trained on ImageNet[33]) Siamese network to explore different feature extractions that can be adapted to various computational capacities. The second step is to include the manually labeled samples in a few-shot setup where the trained model is refined to recognize the new species. It is worth noting that the second step aims to cope with the challenge of identifying new species when only a few samples can be retrieved, which is often the case when great efforts are needed to acquire such samples manually in forests. The third step is to add the explainable AI component to the workflow. Each of these three steps is assessed individually. The result of the proposed workflow is the recognition of individual tree canopies of invasive or native species with an explanation for each classification.

#### Data preparation

The tree species' square cutouts (i.e., tiles) are generated to build training data for the Siamese network's classification workflow. The training cutouts are generated from two sources: (1) tiles generated from the UAV orthomosaic using the Netflora object detection workflow[34] and (2) tiles from the Reforestree benchmark dataset[35]. Netflora is a vanilla Machine Learning workflow that detects forest species from aerial imagery. It was applied to the UAV image, and nine species with various numbers of samples were detected. We extracted two of the nine species, resulting in Species 1 and 2 shown in Figure 2. The species were selected based on the number of predicted objects and their visual distinguishability from other species. Species 1 was selected based on its asymmetrical canopy. Species 2 was selected based on its needle-shaped leaves and symmetric canopy feature. The Reforestree dataset is a labeled dataset of tropical forests based on drone and field data. We utilized the labels *banana*, *cacao*, and *fruit* from this dataset, respectively, resulting in Species 3, 4, and 5 shown in Figure 2. This open dataset also labels other vegetation classes (e.g., citrus and timber) not included in the cutouts. Citrus and timber are excluded because they have fewer labeled samples than the other classes, and including them would lead to an unbalanced dataset and difficulties in training the model.

Data cleaning is applied to all selected cutouts to ensure accuracy and relevance. During the process, the cutouts that display multiple tree canopies of different species within a single cutout are manually detected and removed. The Reforestree dataset had an original spatial resolution of 1 cm, whereas the UAV image of our study has a spatial resolution of 6 cm. This was homogenized by resampling the cutouts from the Reforestree dataset to 6 cm to match the resolutions from both datasets.

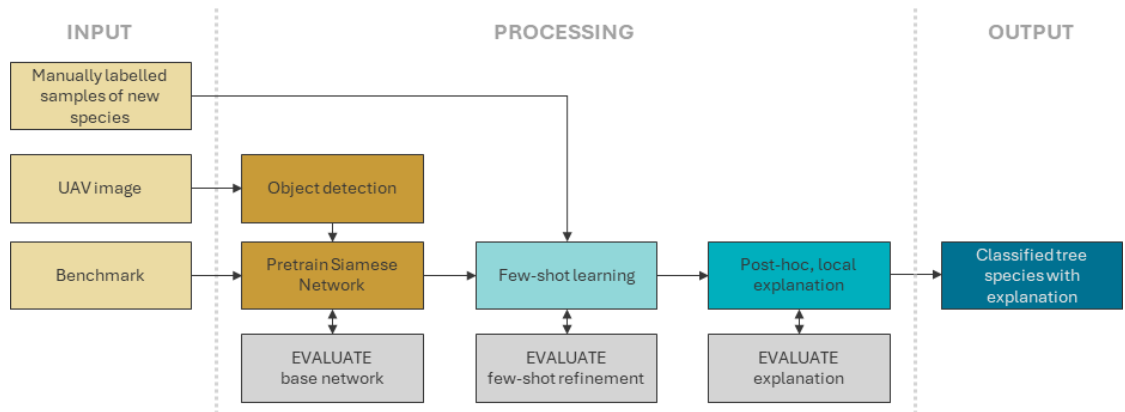

**Figure 1.** Overall workflow of the study, showing the three main processing steps: (1) the pre-training of a base Siamese network using images of tree canopies from existing benchmarks and automatically generated from the UAV image of the study area, (2) refining the networks in a few-shot learning setup, and (3) adding a post-hoc explanation to visually inform users whether the network can distinguish the new tree species.

The size of the cutouts is also cut or padded to a uniform 128 × 128 with zero-padded pixel values. Examples of selected and cleaned cutouts are shown in Figure 2, where one example cutout of each of the five selected species is visualized. Table 1 presents the final number of selected cutout samples.

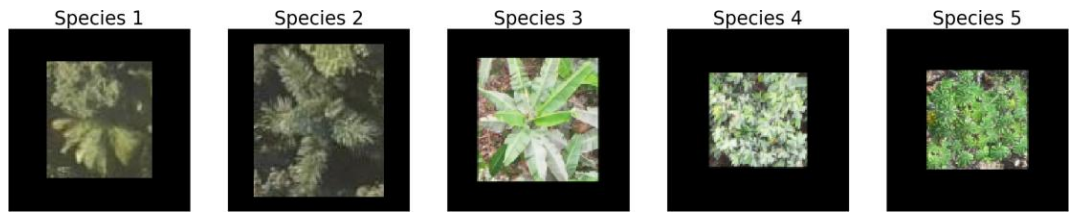

**Figure 2.** Example cutouts generated from the UAV image and Reforestree dataset. One example cutout from each species is shown. Species 1 and 2 are generated using the Netflora workflow from the UAV image of the study area. Species 3, 4, and 5 are banana, cacao, and fruit, respectively, from the Reforesttree dataset.

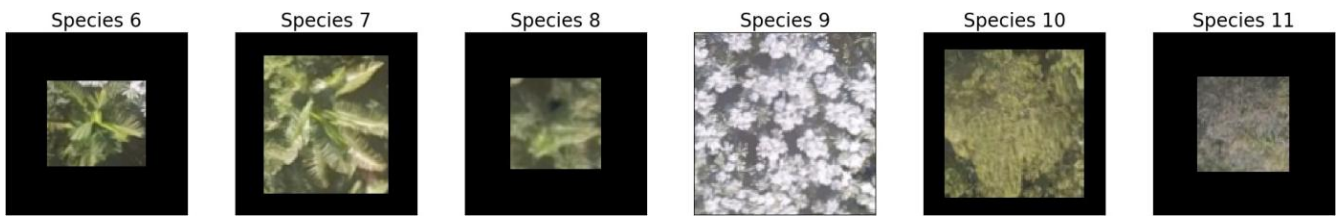

**Figure 3.** Example cutouts for the refinement in the few-shot learning setup generated from the UAV image and Reforestree datasets. All species are identified by experts from the UAV orthomosaic.

### Training the shallow and deep Siamese networks

Similar and dissimilar pairs of cutouts from Species 1 to 5 are generated to train the Siamese Networks. To ensure the model's accuracy, this process considers data balancing in two aspects: 1) the balance in the number of samples between the species and 2) the balanced number of samples between similar and dissimilar pairs. In each of the five classes, 13 samples are randomly selected as candidate cutouts if there are more than 13 samples. To increase the number of candidate cutouts, we apply the following data augmentation methods to each cutout: (i) randomly rotate arbitrary degrees, (ii) horizontal flipping, (iii) vertical flipping, (iv) randomly rotate the image and add a Gaussian noise ($\mu = 0$, $\sigma = 25$) to all RGB channels, and (v) randomly crop the cutout. Together with the original cutouts, the dissimilar pairs are created by exhaustively pairing candidate cutouts between classes, resulting in 2810 dissimilar pairs. We randomly select 10,000 out of 82,810 dissimilar pairs as training data. Similar pairs are created by pairing candidate cutouts within the same class. This results in 04,75 similar pairs. We randomly select 10,000 out of 20,475 similar pairs as training data. The random selections of both similar and dissimilar pairs are performed by generating a uniformly distributed index within each candidate dataset, assuring a non-biased selection. Ultimately, we acquired a balanced training dataset that includes 20,000 sample pairs.

**Table 1.** Number of cutout samples per selected species.

| #Samples in the data for training the base models | | | | | |
|---|---|---|---|---|---|
| Species 1 | Species 2 | Species 3 | Species 4 | Species 5 | |
| 13 | 29 | 26 | 14 | 17 | |
| #Samples in few-shot learning data | | | | | |
| Species 6 | Species 7 | Species 8 | Species 9 | Species 10 | Species 11 |
| 6 | 14 | 30 | 16 | 10 | 6 |

The same training dataset was used to pre-train the shallow CNN-based and MobileNet-based Siamese neural networks, called the base models. The shallow CNN-based Siamese network is initiated with two identical feature extractors, each with four convolution layers. The outputs of the feature extractors are compared with an Euclidean distance layer, resulting in a similarity score between 0 (most dissimilar) and 1 (most similar). We substitute the four convolution layers using the MobieNet backbone trained on the ImageNet dataset[33] to compare with a more expressive feature extractor. The shallow CNN-based and the MobileNet-based Siamese networks have 393K and 3,502K trainable parameters, respectively. Even though the MobileNet-based Siamese network is much more extensive than the shallow CNN-based Siamese network, both networks are lightweight. They can be deployed on devices with limited computational capacity.

The similarity scores between the query object and each support set cutout will be obtained through the Siamese networks, thus resulting in multiple similarity scores between the tree canopy in question and the support set. We propose two methods to predict the species of the query cutout based on the predicted similarity scores. Method 1 classifies the object as the support set species with the highest average similarity score. Method 2 applies the $K$-Neareast-Neighbor ($K$-NN) method, where the $K$ highest similarity scores are selected, and the majority class amongst these $K$ most similar samples will be the predicted label.

*Few-shot learning*
The next step is to use unseen samples for the few-shot learning task where the Siamese networks are refined by new classes representing the target invasive and native species from the study area in São Paulo. This dataset comprises six species; see Table 1 and Figure 2. These cutouts were manually labeled by the experts in Brazil. Similar to the data preparation for training the based models, the cutouts were padded or cut to the same size as the ones in the training dataset above. A maximum of six samples were randomly selected from the six species, and the same augmentation methods were applied to generate balanced similar and dissimilar pairs to refine the based models. To reduce the number of support samples as much as possible, we refined the base models using up to three of the six samples in each class. In order words, we tested the refinement from 1-shot to 3-shot learning. Due to the limited number of remaining samples for testing, we partitioned this dataset using an n-fold cross-validation setting to reduce the random effect on performance evaluation. Namely, the support and test samples were iteratively alternated for each experiment run, and we report the average performance in the Results Section.

*Explanation*
Once the classification is performed, the object will be compared to the support set again by visualizing the highest similarity scores per support set class. If the classes with higher similarity scores also appear more visually similar, the user will have more confidence that the classification is correct. If the network has not learned to consistently identify specific classes, this will likely be evident in the list of similar examples. A standard limitation of XAI studies is the lack of evaluation of the quality of the explanation itself[17]. We utilize the Co-12 concepts proposed in[32] to evaluate the explanation provided. The Co-12 explanation of quality properties is divided into properties regarding the content, presentation, and user dimensions. Here, we will focus on evaluating the content. Evaluating the presentation and user dimensions would require extensive additional experiments with the XAI method's intended end-users, which we will leave to further research. The six Co-12 properties related to the content of the explanation are: *correctness, completeness, consistency, continuity, contrastivity,* and *covariate complexity*. Correctness refers to how faithful the explanation is compared to the workings of the Machine Learning model. Here, the correctness will be quantified by counting how often the result of the explanation (i.e., the class that would be assigned if only taking the majority of the most similar samples) is the same as the classification assigned through the model. Completeness refers to how much of the Machine Learning method is described by the explanation. In our case, the workflow consists of object detection, a Siamese Network to determine similarity and a classification assigned based on these similarity measures. The explanation only considers the final step, where the class label is assigned based on the similarity metrics calculated by the Siamese networks. This limits the completeness of the explanation but heightens its correctness. Consistency refers to the deterministicness of the explanation. That is to say, is the explanation sensitive to changes if the model is run multiple times? In our case, the weights of the trained model are static, so the similarity metrics will not change if a query pair is run through it multiple times. Given the same support set of images, the explanation will be consistent. Continuity refers to how stable the explanation is to small perturbations in the input data. This will be tested by slightly changing an input image and verifying whether the explanation changes. Contrastivity refers to the variety of explanations created for different samples. This will be evaluated by quantifying the diversity of the selection of support images for the explanation. Covariate complexity refers to the complexity of an explanation of semantic meaning. Quantitatively measuring this complexity would involve user studies, which are not included due to the same reasoning as the presentation and user dimension Co-12 concepts.

**Experimental setup and evaluation metrics**

The proposed workflow is tested through three sets of experiments:

1. **From Siamese Networks to multi-class classification for the base models** – This set of experiments aims to establish a baseline of classification performance before the few-shot learning step. It uses the training dataset in Table 1 to compare whether better results are obtained when utilizing the average similarity score versus $K$-NN to convert the similarity scores to a classification label, as well as comparing the accuracy that can be achieved by the shallow CNN-based versus MobileNet-based Siamese networks. The classification performance is measured in standard metrics, including Precision, Recall, F1-score, and Accuracy.

2. **Few-shot learning for the refinement of the base models** – The second set of experiments aims to test the accuracy of the few-shot learning task for recognizing new species by refining the base models. Experiments are performed to evaluate the $n$-fold $k$-shot classification. Namely, we iteratively alternated the partitioning of support and test samples for each independent training of models to avoid random effects on evaluation, and we varied the number of support samples per class, i.e., shots ($k$) from 0 to 3. Results are presented for the refined shallow CNN-based and MobileNet-based Siamese Networks using the optimum base models determined in the first set of experiments to assign class labels.

3. **Evaluation of explanation** - Using the Siamese network, which obtains the best performance, example local explanations are given for the prediction of the unseen classes. The correctness $C_{cor}$, continuity $C_{cty}$, and contrastivity $C_{cst}$ of the explanation is quantified by the following equations. Let $\{I_1, I_2, ..., I_N\}$ represent the set of test image samples which has a true class $Y(I_n)$, a class predicted by the model $\hat{Y}(I_n)$, and the support set of images $S = \{S_1, S_2, ..., S_K\}$ selected for the explanation, where $N$ and $K$ stand for the total number of test samples and support samples in each class, respectively. The correctness is quantified by comparing the number of samples in the explanation that have the class predicted by the model as follows:

$$\delta(n) = \begin{cases} 1 & \text{if } \hat{Y}(I_n) = Y(S_k) \, \& \, \hat{Y}(I_n) = Y(I_n), \\ 0 & \text{otherwise.} \end{cases} \quad (1)$$

$$C_{cor} = \frac{1}{N}\frac{1}{K} \sum_{n=1}^{N} \sum_{k=1}^{K} \delta(n). \quad (2)$$

The continuity considers whether the explanation changes with slight variations in the input data. Here, we apply a random data augmentation method as described above to a test sample image $I_n$ to generate the modified image $I'$. We then see whether the images selected for the explanation are changed and calculate the proportion of explanation images that have remained unchanged for all samples as follows:

$$\gamma(n) = \begin{cases} 1 & \text{if } S_i \in S' \, \& \, S, \\ 0 & \text{otherwise.} \end{cases} \quad (3)$$

$$C_{cty} = \frac{1}{N}\frac{1}{K} \sum_{n=1}^{N} \sum_{i=1}^{K} \gamma(n). \quad (4)$$

The contrastivity considers the diversity in the images selected for the explanations and is calculated as follows:

$$C_{cst} = \frac{1}{N} \sum_{n=1}^{N} \left( -\sum_{k=1}^{K} p(S_k) \log_2 p(S_k) \right) / \log_2(K), \quad (5)$$

where $C_{cst}$ is the Shannon Entropy representing the diversity in the selection of support images selected for the explanation, $p(S_k)$ is the probability of selecting the support set image $S_k$ for an explanation across all the samples, and $K$ is the total number of images selected for an explanation. The Shannon Entropy is normalized by the maximum entropy value to scale it from 0 to 1, making the contrastivity value independent of the number of sample explanations.

## Results and Discussion

### From Siamese Networks to multi-class classification

The output of the Siamese network is a list of similarity scores for an input test image and images from a support set. This first set of experiments analyzed the sensitivity of the classification results to (1) using average accuracy or $K$-NN to assign the class and (2) the performance of the shallow CNN-based versus MobileNet-based Siamese Networks. Results for the base models are presented in Table 2. Generally, both base models achieved satisfactory classification performance, e.g., with a Precision reaching 1.00. This is because an adequate number of samples, i.e., 13 in each class, was used to train the models to match the test and support sample images. However, the MobileNet-based Siamese model achieves much higher Recall, F1-score, and Accuracy than that of the shallow CNN-based Siamese model, indicating a much more robust feature extractor pre-trained on the ImageNet dataset[33]. Interestingly, as the models already reached a stable, high-performance level based on very few support samples, e.g., $k = 1$, the average and different $K$-NN metrics to assign the class did not lead to a noticeable performance difference.

**Table 2.** Classification performance for the shallow and deep base models. The performance is tested on Species 2 to 4 samples that are not used in base model training. Species 1 is excluded since all samples are exhausted in training. Three measures are used to evaluate the performance: Precision, Recall, and F1-score. An average Accuracy weighted by the number of samples is computed per classification method.

| Network | Shallow CNN | | | | MobileNet | | | |
|---|---|---|---|---|---|---|---|---|
| Metric | Precision | Recall | F1-score | Accuracy | Precision | Recall | F1-score | Accuracy |
| Avg. | 0.98 | 0.83 | 0.89 | 0.82 | 1.00 | 0.94 | 0.97 | 0.97 |
| KNN ($k$=1) | 0.98 | 0.82 | 0.89 | 0.82 | 1.00 | 0.97 | 0.99 | 0.97 |
| KNN ($k$=2) | 0.98 | 0.82 | 0.89 | 0.82 | 1.00 | 0.97 | 0.99 | 0.97 |
| KNN ($k$=3) | 0.98 | 0.82 | 0.89 | 0.82 | 1.00 | 0.94 | 0.97 | 0.94 |

### Few-shot learning

The second set of experiments analyzed the performance of the few-shot learning, refining the base models determined from the first set of experiments using a few support samples. To more intuitively show the benefits of refining the base models using increasing support samples ($k = 1, 2, 3$) for classifying new species, the performance differences for both the shallow CNN- and MobileNet-based models are compared before and after the refinement.

The classification results are measured in Precision (Figure 4a), Recall (Figure 4b), F1-score (Figure 4c), and Accuracy (Figure 4d). First, with few-shot learning, both the shallow CNN- and MobileNet models performed better after being refined using a few support samples than the zero-shot learning (i.e., before the refinement), and the more support samples, the better the performance. Second, the MobildeNet-based Siamese network performs consistently better than the shallow CNN-based models in all the classification metrics from 1-shot to 3-shot learning, except for the 1-shot learning measured in Precision. The 3-shot learning from the MobileNet-based Siamese network achieved the best classification performance, e.g., Precision 0.90, F1-score 0.86, underscoring the effectiveness of using a few support samples to detect newly invasive tree species. Last, it is interesting to observe that the performance gap between the shallow CNN- and MobileNet-based Siamese networks is more profound for 3-shot learning than 1- or 2-shot learning. As the MobileNet engages more trainable parameters than the shallow CNN (3,502K versus 393K), the MobileNet feature extractor becomes more effective when more support samples are leveraged to refine the model.

### Explanation

In addition to the classification performance, the evaluation of the results' explanation is measured in correctness ($C_{cor}$), continuity ($C_{cty}$), and contrastivity ($C_{cst}$), as reported in Table 3. Because only one support sample image exists in the 1-shot learning, $C_{cst}$ cannot be measured. The correctness and continuity are similar to the classification results for the shallow CNN- and MobileNet-based Siamese networks. Both models demonstrated a very high contrastivity for the support samples.

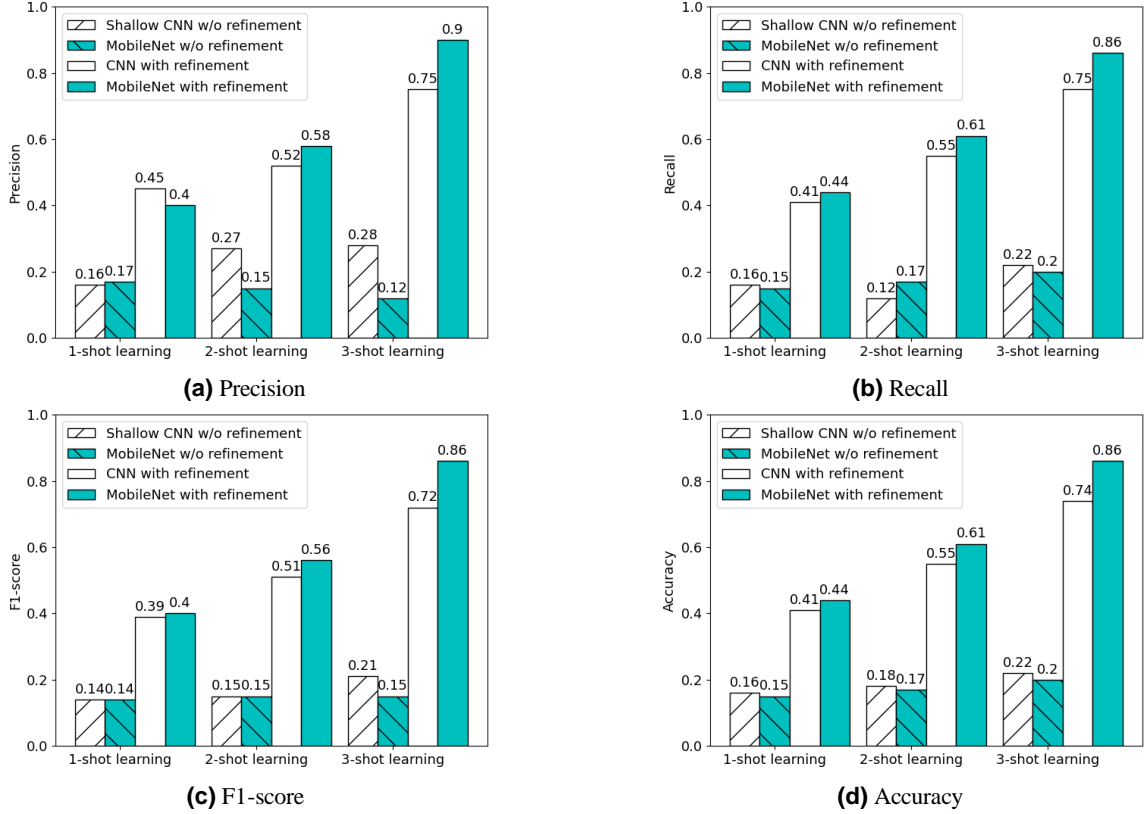

**Figure 4.** The classification results of the shallow CNN- and MobileNet-based Siamese models without and with the refinement for few-shot learning.

**Table 3.** Evaluation of the explanation for n-shot learning using the shallow CNN- and MobileNet-based siamese networks before and after the refinement.

| Network | Shallow CNN | | | | | | MobileNet | | | | | |
| --- | --- | --- | --- | --- | --- | --- | --- | --- | --- | --- | --- | --- |
| Metric | Before refinement | | | After refinement | | | Before refinement | | | After refinement | | |
| few-shot | $C_{cor}$ | $C_{cty}$ | $C_{cst}$ | $C_{cor}$ | $C_{cty}$ | $C_{cst}$ | $C_{cor}$ | $C_{cty}$ | $C_{cst}$ | $C_{cor}$ | $C_{cty}$ | $C_{cst}$ |
| $k=1$ | 0.16 | 0.76 | - | 0.41 | 0.89 | - | 0.15 | 0.64 | - | 0.45 | 0.83 | - |
| $k=2$ | 0.06 | 0.80 | 1.00 | 0.51 | 0.84 | 0.98 | 0.18 | 0.73 | 1.00 | 0.54 | 0.70 | 0.99 |
| $k=3$ | 0.10 | 0.81 | 1.00 | 0.51 | 0.87 | 0.96 | 0.06 | 0.75 | 1.00 | 0.60 | 0.81 | 0.98 |

Moreover, Figure 5 shows the qualitative results of the 3-shot learning for the new tree species. In addition to the similarity score, the classification results become more interpretable thanks to example explanations and associated explanation metrics, i.e., correctness, continuity, and contrastivity. For example, the correctness of prediction results for the input Species 7 is 0.67 because the third support sample image has a very low similarity score (0.07) to the input, which was a false positive support. Another interesting example is the third support sample for the input Species 10. One can quickly notice that the third support sample is more visually different than the other support samples to the input sample, even though they belong to the same class. Consequently, its similarity score is much lower than the other support samples.

## Conclusions

This study uses drone-captured RGB images to introduce a simple yet effective Siamese-based method for tree species recognition in São Paulo's Atlantic Forest. Using few-shot learning minimizes the need for extensive labeled data, making it practical for real-world applications. Two shallow CNN-based and a MobileNet-based Siamese networks were proposed, with the latter achieving a 0.90 Precision in 3-shot learning. Though larger, the MobileNet-based network is still lightweight and suitable for mobile deployment. A case-based strategy improves explainability with visual examples and metrics, opening new opportunities for AI and UAVs in forest management and biodiversity conservation.

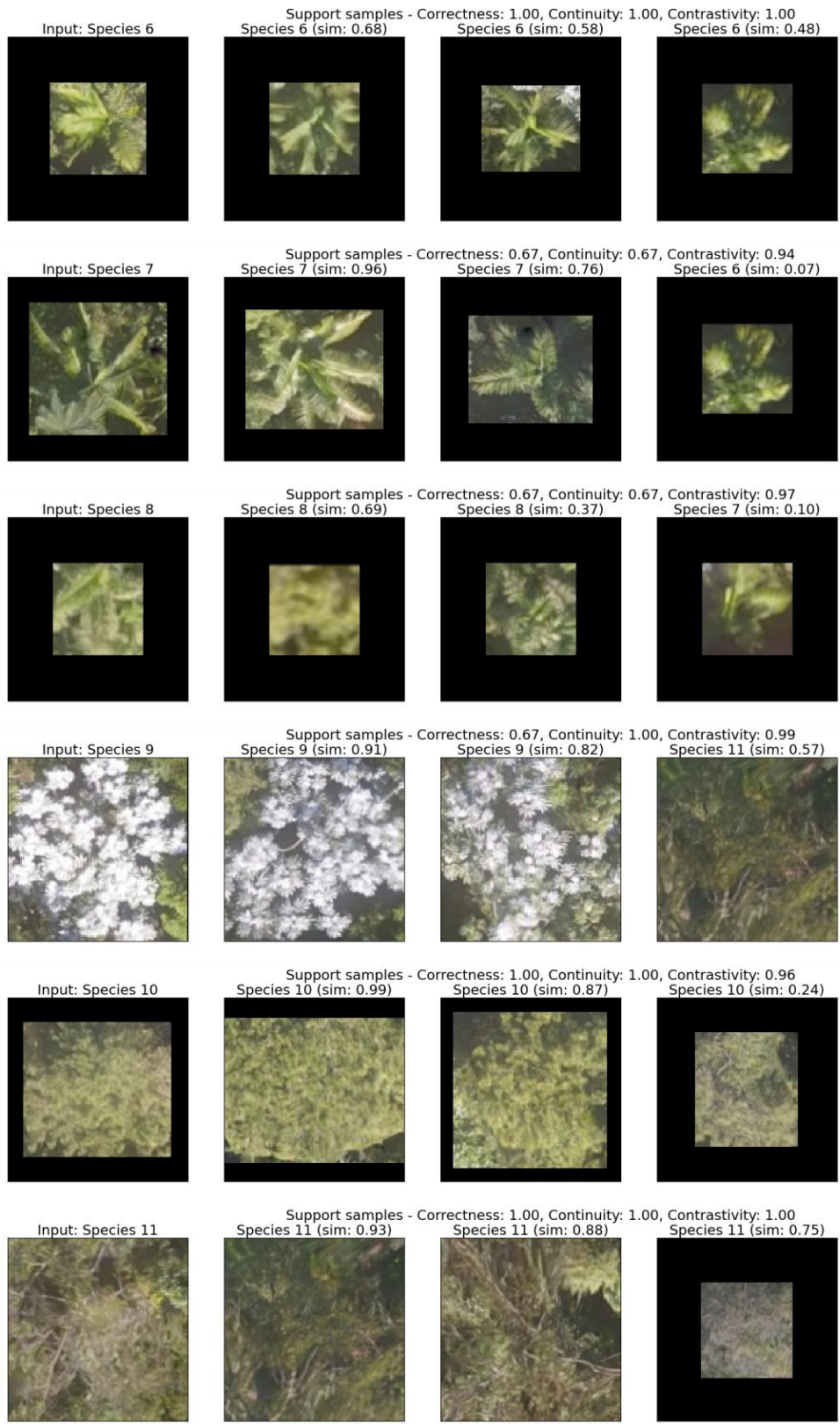

**Figure 5.** Example explanations of the prediction results for 3-shot learning. The left-most column shows the input image and the other columns show the support sample images, and each row represents a different species. In addition to the similarity score (sim), the explanation of the visual results measured in Correctness, Continuity, and Contrastivity is on top of each row.

## Data Availability

The training datasets for the experiments are available on Zenodo, [PERSISTENT WEB LINK TO DATASETS WILL BE GIVEN AFTER ACCEPTANCE] and all code to reproduce the experiments is available on an open GitHub repository, [PERSISTENT WEB LINK TO CODE WILL BE GIVEN AFTER ACCEPTANCE].

## Acknowledgements

This publication is part of the project "Bridging the gap between artificial intelligence and society: Developing responsible and viable solutions for geospatial data" (with project no. 18091) of the research program Talent Program Veni 2020, which is (partly) financed by the Dutch Research Council (NWO). The authors would also like to thank the São Paulo Municipal Green and Environment Secretariate for the procurement and labeling of the UAV data and in-situ labels, and the teams behind the Reforestree and Netflora projects for sharing their code and data online.

## Author contributions statement

C.G, P.C., and A.A.P. contributed to the conceptualization of the ideas. A.A.P. supported the data curation. H.C., O.K., S.G., F.N., and P.C contributed to the development of the methods and execution of the experiments. C.G., F.D.J., H.C., O.K., S.G., and A.A.P. contributed to the writing of the manuscript.

## Competing interests

The authors declare no competing interests.